\documentclass{article}

\usepackage{PRIMEarxiv}
\usepackage{tabularx}
\usepackage{amsmath}
\usepackage[utf8]{inputenc} 
\usepackage[T1]{fontenc}    
\usepackage{hyperref}       
\usepackage{url}            
\usepackage{booktabs}       
\usepackage{amsfonts}       
\usepackage{nicefrac}       
\usepackage{microtype}      
\usepackage{lipsum}
\usepackage{fancyhdr}       
\usepackage{graphicx}       
\graphicspath{{media/}}     

\title{Reducing False Ventricular Tachycardia Alarms in
ICU Settings: A Machine Learning Approach}

\author{
  Grace Funmilayo Farayola \\ 
  University of Buckingham, \\ Buckingham, UK 
  \And
  Akinyemi Sadeeq Akintola \\ 
  Universidade NOVA de Lisboa, \\Lisbon, Portugal 
  \And
  Oluwole Fagbohun \\ 
  Readrly Limited, London,\\ UK 
  \AND
  Chukwuka Michael \\ 
  Readrly Limited, London,\\ UK 
  \And
  Bisola Faith Kayode \\ 
  Independent Researcher,\\ London, UK 
  \And
  Christian Chimezie \\ 
  Independent Researcher,\\ Bristol, UK 
  \AND
  Temitope Kadri \\ 
  Readrly Limited, London,\\ UK 
  \And
  Abiola Oludotun \\ 
  Readrly Limited, London,\\ UK 
  \And
  Nelson Ogbeide \\ 
  Independent Researcher,\\ London, UK 
  \AND
  Mgbame Michael \\ 
  Hankali Intel, Lagos,\\ Nigeria 
  \And
  Adeseye Ifaturoti \\ 
  University of Greenwich,\\ London, UK 
  \And
  Toyese Oloyede \\ 
  Independent Researcher, \\Northampton, UK 
}

\begin{document}
\maketitle

\begin{abstract}
False arrhythmia alarms in intensive care units (ICUs) are a significant challenge, contributing to alarm fatigue and potentially compromising patient safety. Ventricular tachycardia (VT) alarms are particularly difficult to detect accurately due to their complex nature. This paper presents a machine learning approach to reduce false VT alarms using the VTaC dataset, a benchmark dataset of annotated VT alarms from ICU monitors. We extract time-domain and frequency-domain features from waveform data, preprocess the data, and train deep learning models to classify true and false VT alarms. Our results demonstrate high performance, with ROC-AUC scores exceeding 0.96 across various training configurations. This work highlights the potential of machine learning to improve the accuracy of VT alarm detection in clinical settings.
\end{abstract}


\vspace{1\baselineskip}
\hrule
\vspace{1\baselineskip}
\textbf{NOTE: Preprint, Accepted to ICMLT 2025, Helsinki, Finland}  

\vspace{0.2\baselineskip}

Grace Funmilayo Farayola is with the University of Buckingham, Buckingham, UK (e-mail: gracefarayola@gmail.com). Akinyemi Sadeeq Akintola is with Universidade NOVA de Lisboa, Lisbon, Portugal (e-mail: Sadeeq2@gmail.com).  
Oluwole Fagbohun is with Readrly Limited, London, UK (e-mails: \{wole@readrly.io).  
Bisola Faith Kayode and Nelson Ogbeide are Independent Researchers based in London, UK (e-mails: \{bisolakayode11, ogbeide331\}@gmail.com).  
Christian Chimezie is an Independent Researcher based in Bristol, UK (e-mail: Chimeziechristiancc@gmail.com).  
Temitope Kadri, Chukwuka Michael Oforgu and Abiola Oludotun are with Readrly Limited, London, UK (e-mails: \{temitopekadri, Dotunodude, chukwuka.oforgu\}@gmail.com).  
Mgbame Michael is with Hankali Intel, Lagos, Nigeria (e-mail: michaelmgbame@yahoo.com).  
Adeseye Ifaturoti is with the University of Greenwich, London, UK (e-mail: Adeseye31@gmail.com).  
Toyese Oloyede is an Independent Researcher based in Northampton, UK (e-mail: Toyesej@gmail.com).  

\vspace{0.2\baselineskip}
\hrule

\section{Introduction}
False arrhythmia alarms in ICUs are a persistent problem, with studies indicating that a significant proportion of alarms are false positives \cite{nguyen2020double}. These false alarms contribute to alarm fatigue among healthcare providers, reducing their responsiveness to critical alerts and potentially endangering patient safety \cite{clifford2015physionet}. Among arrhythmia alarms, ventricular tachycardia (VT) \cite{koplan2009ventricular} is particularly challenging to detect accurately due to its complex waveform characteristics and the need for high sensitivity and specificity \cite{akhtar1990clinical}.

The problem of false alarms in ICUs has been widely documented in the literature. For instance, studies by \cite{muroi2020automated} have highlighted that up to 90\% of ICU alarms are false positives, leading to desensitization among healthcare providers. This phenomenon, known as alarm fatigue, has been linked to delayed responses to true alarms \cite{fernandes2020detecting} and adverse patient outcomes. The challenge is further compounded by the variability in ECG waveforms and the presence of noise and artifacts in ICU monitoring systems \cite{goodwin2023high}.

Recent advancements in machine learning and signal processing have shown promise in addressing this issue. For example, \cite{zhou2022contrastive} demonstrated the effectiveness of deep learning models in reducing false arrhythmia alarms by leveraging large datasets of annotated ECG signals \cite{zhou2022contrastive}. Similarly, \cite{pflugradt2017fast} proposed a multi-modal approach combining ECG and photoplethysmogram (PPG) data to improve the accuracy of VT detection. However, the lack of large, diverse, and annotated datasets has limited the generalizability of these approaches \cite{ali2023multi}.

To address this gap, we utilize the VTaC dataset \cite{lehman2024vtac}, a benchmark dataset of over 5,000 annotated VT alarms from ICU monitors. This dataset includes waveform data from multiple hospitals and manufacturers, ensuring diversity and representativeness. The availability of high-quality annotations and the inclusion of multiple physiological signals (e.g., ECG, PPG, and arterial blood pressure) make the VTaC dataset an ideal resource for developing and evaluating machine learning models for VT alarm classification.

\begin{figure}[h]
\centering
\includegraphics[scale=0.6]{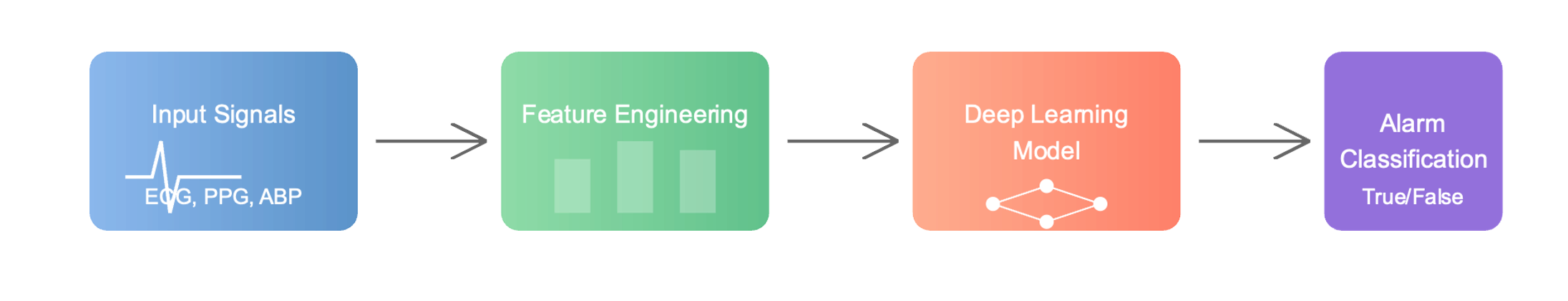}
\caption{Pipeline for false arrhythmia alarm detection in ICU settings. The system processes multiple physiological signals through feature engineering before classification using deep learning models.}
\label{fig1}
\end{figure}

In this paper, we present a comprehensive machine learning pipeline for reducing false VT alarms. Our approach involves preprocessing the waveform data, extracting relevant features, and training deep learning models to classify true and false alarms. Figure \ref{fig1} depict the whole pipeline of methodology. 

\section{Methodology}
\begin{figure*}[htbp]
\centering
\includegraphics[scale=0.5]{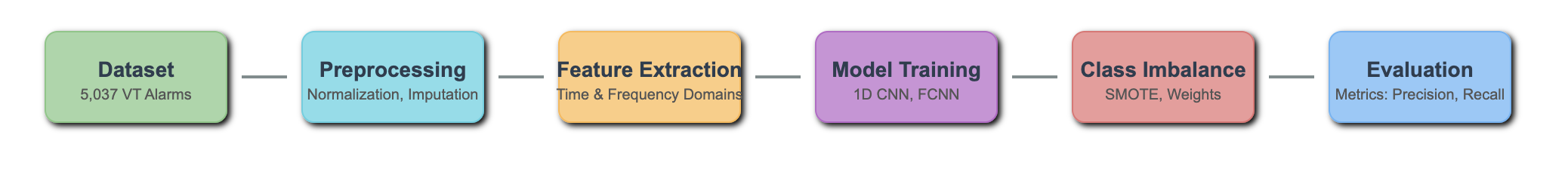}
\caption{Pipeline for VT Alarm Classification.}
\label{fig2}
\end{figure*}

The methodology employed in this study is designed to address the challenge of reducing false ventricular tachycardia (VT) alarms in intensive care unit (ICU) settings. The approach involves a systematic pipeline that includes data preprocessing, feature extraction, and the development of deep learning models tailored to handle the unique characteristics of physiological waveform data. The preprocessing stage ensures the data is clean and normalized, while feature extraction captures both time-domain and frequency-domain characteristics of the signals. Two deep learning architectures—a 1D convolutional neural network (CNN) with multi-head attention and a fully connected neural network (FCNN)—are proposed to classify VT alarms effectively. Additionally, strategies such as Synthetic Minority Over-sampling Technique (SMOTE) and class weighting are implemented to mitigate the class imbalance inherent in the dataset. This comprehensive methodology aims to improve the accuracy and generalizability of VT alarm classification systems. Figure \ref{fig2} illustrates the pipeline, which begins with dataset collection, followed by preprocessing, feature extraction, model training, class imbalance handling, and evaluation. Each step is color-coded for clarity, with distinct colors representing different stages of the pipeline.

\subsection{Dataset}
The VTaC dataset is a publicly available benchmark dataset designed for the development and evaluation of algorithms to reduce false VT alarms in ICU settings. It consists of 5,037 annotated VT alarm events, including 1,441 true alarms and 3,596 false alarms. Each waveform recording contains at least two ECG leads and one or more pulsatile waveforms, such as photoplethysmogram (PPG) and arterial blood pressure (ABP). The dataset is derived from ICU monitors across three major U.S. hospitals, ensuring a diverse and representative collection of waveform data.

Each waveform recording spans a 6-minute segment, capturing 5 minutes of data before the alarm onset and 1 minute after. This temporal context is critical for analyzing the physiological changes leading up to the alarm. The dataset was annotated by at least two independent experts, with conflicts resolved through adjudication, ensuring high-quality labels for training and evaluation. The annotations include detailed information about the alarm type, signal quality, and clinical context, making the dataset suitable for both supervised and unsupervised learning tasks 
The dataset is stored in WFDB format, a standard format for physiological waveform data. Each patient record contains up to five VT alarm events, ensuring a balanced representation of alarms across patients. The dataset also includes a predefined benchmark split for training, validation, and testing, facilitating reproducible research. This split ensures that the model is evaluated on unseen data, providing a robust assessment of its performance.

One of the key strengths of the VTaC dataset is its diversity. The dataset includes waveform data from multiple manufacturers of ICU monitoring systems, as well as data from patients with a wide range of clinical conditions. This diversity enhances the generalizability of machine learning models trained on the dataset, making them more likely to perform well in real-world clinical settings. Additionally, the dataset includes metadata such as patient demographics, clinical history, and medication information, which can be used to further enhance model performance.

Despite its strengths, the VTaC dataset has some limitations. For example, the dataset is relatively small compared to other benchmark datasets in the field of machine learning. This limits the complexity of models that can be trained effectively on the dataset. Additionally, the dataset is imbalanced, with false alarms significantly outnumbering true alarms. This imbalance must be addressed during model training to avoid bias toward the majority class. Overall, the VTaC dataset represents a valuable resource for researchers and practitioners working on the problem of false VT alarms in ICU settings.

\subsection{Preprocessing}
The preprocessing pipeline begins with loading the waveform data from .dat and .hea files using the wfdb library. This step ensures that the raw waveform data is accessible for further processing. Missing values, which can arise due to signal artifacts or recording errors, are handled using mean imputation. This approach replaces missing values with the mean of the corresponding feature, ensuring that the dataset remains complete.

Normalization is a critical step in preprocessing, as it ensures that all features are on a comparable scale. We apply Min-Max scaling, which transforms the data to a range of [0, 1]. This step is particularly important for deep learning models, as it helps stabilize training and improve convergence. Finally, the dataset is split into training, validation, and test sets, with 80\% of the data used for training, 10\% for validation, and 10\% for testing. This split ensures that the model is evaluated on unseen data, providing a robust assessment of its performance.

\subsection{Feature Extraction}
Feature extraction is a crucial step in transforming raw waveform data into meaningful inputs for machine learning models. We extract both time-domain and frequency-domain features to capture the essential characteristics of the waveforms. Time-domain features include mean, standard deviation, skewness, kurtosis, and root mean square (RMS). These features provide insights into the amplitude and variability of the signals.

Frequency-domain features are extracted using Welch’s method for power spectral density estimation. These features include dominant frequency, spectral entropy, and mean coherence between channels. Additionally, we compute wavelet energy using the continuous wavelet transform (CWT) with a Morlet wavelet. These features capture the frequency content and temporal dynamics of the signals, enabling the model to distinguish between true and false alarms.

The extracted features are concatenated into a single feature vector for each waveform recording. This vector serves as the input to the machine learning models. By combining time-domain and frequency-domain features, we ensure that the models have access to a comprehensive representation of the waveform data.

\subsection{Model Architecture}

We propose two deep learning architectures for this task: a 1D convolutional neural network (CNN) with multi-head attention and a fully connected neural network (FCNN). Both models are designed to handle the unique challenges of VT alarm classification, including high-dimensional input data and class imbalance.

\begin{figure}[htbp]
\centering
{\includegraphics[scale=0.45]{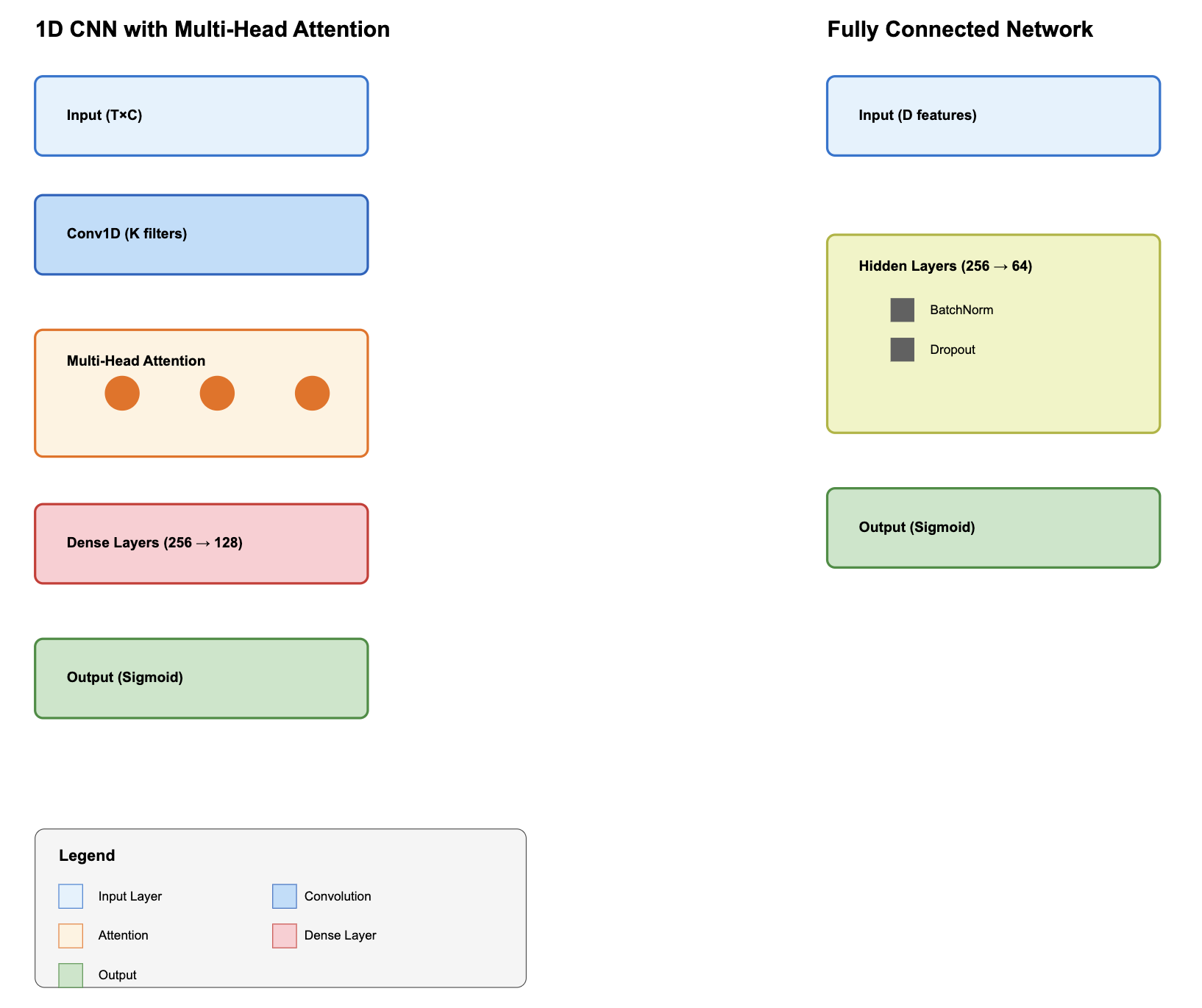}}
\caption{Architectural overview of the proposed models: (Left) A 1D Convolutional Neural Network (CNN) with Multi-Head Attention for processing time-series signals, comprising convolutional layers, attention mechanisms, and fully connected layers. (Right) A Fully Connected Neural Network (FCNN) for feature-based classification, featuring batch normalization and dropout regularization. Layer types and dimensions are annotated, with a legend clarifying component representations.}
\label{fig3}
\end{figure}

As illustrated in Figure \ref{fig3}, the dual architecture framework includes (1) a 1D CNN with multi-head attention for raw waveform analysis, and (2) an FCNN for engineered feature processing. Both models culminate in sigmoid-activated outputs for binary classification, with detailed layer configurations and regularization strategies explicitly visualized.

\subsection{1D Convolutional Neural Network with Multi-Head Attention}
The 1D CNN processes raw waveform data directly, leveraging convolutional layers to extract local patterns and multihead attention to capture long-range dependencies. The input to the model is a time-series signal ( $\boldsymbol{X} \in R^{T \times C}$ ) where $(T)$ is the number of time steps and $(C)$ is the number of channels (e.g., ECG leads, PPG, ABP).

The first layer applies a 1D convolution with ( $K$ ) filters of size $(F)$, followed by batch normalization and a ReLU activation function:

\begin{equation}
\boldsymbol{H}_{\mathbf{1}}=\operatorname{ReLU}(\operatorname{BatchNorm}(\operatorname{Conv} 1 D(\boldsymbol{X}, K, F))) .
\end{equation}

The output ($\boldsymbol{H}_{\mathbf{1}} \in R^{T^{\prime} \times K}$) is passed through a max-pooling layer to reduce dimensionality: $\boldsymbol{H}_{\mathbf{2}}=\operatorname{MaxPoollD}\left(\boldsymbol{H}_{\mathbf{1}}\right)$. Next, the multi-head attention mechanism computes attention scores for each time step. For ($H$) attention heads, the attention output is given by: \\
$\boldsymbol{H}_{3}=\operatorname{MultiHeadAttention}\left(\boldsymbol{H}_{2}, \boldsymbol{H}_{2}, \boldsymbol{H}_{2}\right)$, where the attention mechanism is defined as: \\
Attention $(\boldsymbol{Q}, \boldsymbol{K}, \boldsymbol{V})=\operatorname{Softmax}\left(\frac{\rho \boldsymbol{K}^{T}}{\sqrt{d_{k}}}\right) \boldsymbol{V}$. Here, $(\boldsymbol{Q}),(\boldsymbol{K})$, and $(\boldsymbol{V})$ are learned linear transformations of $\left(\boldsymbol{H}_{2}\right)$, and $\left(d_{k}\right)$ is the dimensionality of the key vectors. The output $\left(\boldsymbol{H}_{3}\right)$ is passed through a global average pooling layer to obtain a fixed-size representation: $\boldsymbol{h}=$ GlobalAveragePoolinglD $\left(\boldsymbol{H}_{3}\right)$

Finally, the pooled features are passed through two fully connected layers with ReLU activation and dropout for regularization: \\
$\boldsymbol{y}=\sigma\left(\boldsymbol{W}_{2} \operatorname{RL}\left(\boldsymbol{W}_{\mathbf{1}} \boldsymbol{h}+\boldsymbol{b}_{\mathbf{1}}\right)+\boldsymbol{b}_{\mathbf{2}}\right)$ where $(\sigma)$ is the sigmoid activation function, and $\left(\boldsymbol{W}_{\mathbf{1}}\right),\left(\boldsymbol{W}_{2}\right),\left(\boldsymbol{b}_{\mathbf{1}}\right)$, and $\left(\boldsymbol{b}_{2}\right)$ are learnable parameters.

\subsection{Fully Connected Neural Network}
The FCNN processes the extracted feature vector $\mathcal{A b}\{x\} \in \mathbb{A}\{R\}^{\wedge} D$, where $D$ is the number of features. The model consists of four hidden layers with batch normalization and dropout for regularization. The output of the $(l)$-th layer is given by:  

\[
\boldsymbol{h}_{\boldsymbol{l}}=\operatorname{ReLU}(\operatorname{BatchNorm}(\boldsymbol{W}_{\boldsymbol{l}} \boldsymbol{h}_{\boldsymbol{l}-\mathbf{1}}+\boldsymbol{b}_{\boldsymbol{l}}))
\]

where $\boldsymbol{h}_{\mathbf{0}}=\boldsymbol{x}$, and $\boldsymbol{W}_{\boldsymbol{l}}$ and $\boldsymbol{b}_{\boldsymbol{l}}$ are learnable parameters. Dropout is applied to each hidden layer to prevent overfitting:  

\[
\boldsymbol{h}_{\boldsymbol{l}}=\operatorname{Dropout}(\boldsymbol{h}_{\boldsymbol{l}}, p)
\]

where $p$ is the dropout probability. The final layer uses a sigmoid activation function for binary classification:  

\[
\boldsymbol{y}=\sigma(\boldsymbol{W}_{\boldsymbol{L}} \boldsymbol{h}_{\boldsymbol{L}-\mathbf{1}}+\boldsymbol{b}_{\boldsymbol{L}})
\]

\begin{table}
  \caption{Architecture of the 1D CNN with Multi-Head Attention}
  \centering
  \renewcommand{\arraystretch}{1.2}
  \begin{tabular}{lllc}
    \toprule
    Layer & Description & Output Shape & Parameters \\
    \midrule
    Input & Time-series signal $\mathbf{X} \in \mathbb{R}^{T \times C}$ & $(T,C) \times (T,C)$ & $T$: Time steps, $C$: Channels \\
    1D Convolution & $K$ filters of size $F$, BatchNorm, ReLU & $(T',K) \times (T,K)$ & $K$: Filters, $F$: Filter size \\
    Max Pooling & Reduces dimensionality using pooling & $(T'',K) \times (T'',K)$ & Pool size: 2 \\
    Multi-Head Attention & Attention with $H$ heads & $(T'',K) \times (T'',K)$ & $H$: Heads, $d_k$: Key dim. \\
    Global Avg Pooling & Aggregates features across time steps & $(K) \times (K)$ & - \\
    Fully Connected 1 & Dense + ReLU + Dropout & $(256) \times (256)$ & $W_1, b_1$ \\
    Fully Connected 2 & Dense + ReLU + Dropout & $(128) \times (128)$ & $W_2, b_2$ \\
    Output Layer & Sigmoid activation (binary classification) & $(1) \times (1)$ & $W_3, b_3$ \\
    \bottomrule
  \end{tabular}
  \label{tab:cnn_architecture}
\end{table}

\begin{table}
  \caption{Architecture of the Fully Connected Neural Network (FCNN)}
  \centering
  \renewcommand{\arraystretch}{1.2}
  \begin{tabular}{lllc}
    \toprule
    Layer & Description & Output Shape & Parameters \\
    \midrule
    Input & Extracted feature vector $\mathbf{x} \in \mathbb{R}^{D \times D}$ & $(D) \times (D)$ & $D$: Number of features \\
    Hidden Layer 1 & Dense + ReLU + BatchNorm + Dropout & $(256) \times (256)$ & $W_1, b_1$ \\
    Hidden Layer 2 & Dense + ReLU + BatchNorm + Dropout & $(192) \times (192)$ & $W_2, b_2$ \\
    Hidden Layer 3 & Dense + ReLU + BatchNorm + Dropout & $(128) \times (128)$ & $W_3, b_3$ \\
    Hidden Layer 4 & Dense + ReLU + BatchNorm + Dropout & $(64) \times (64)$ & $W_4, b_4$ \\
    Output Layer & Sigmoid activation (binary classification) & $(1) \times (1)$ & $W_5, b_5$ \\
    \bottomrule
  \end{tabular}
  \label{tab:fcnn_architecture}
\end{table}

\subsection{Handling Class Imbalance}
Class imbalance is a common challenge in medical datasets, where the number of false alarms often far exceeds the number of true alarms. To address this issue, we employ two strategies: Synthetic Minority Over-sampling Technique (SMOTE) and class weights.

SMOTE generates synthetic samples for the minority class by interpolating between existing samples. This approach helps balance the dataset, ensuring that the model is not biased toward the majority class.

Class weights are another effective strategy for handling class imbalance. By assigning higher weights to the minority class during training, we ensure that the model pays more attention to true alarms. This approach is particularly useful when the dataset cannot be resampled due to computational constraints. Both strategies are evaluated in our experiments, with SMOTE achieving slightly better performance.

\begin{table}
  \caption{Strategies for Handling Class Imbalance}
  \centering
  \renewcommand{\arraystretch}{1.2}
  \begin{tabularx}{\textwidth}{p{3cm} p{4cm} p{4cm} p{4cm}}
    \toprule
    Strategy & Description & Advantages & Disadvantages \\
    \midrule
    SMOTE & Generates synthetic samples by interpolating between existing samples & Balances dataset, reduces bias & May introduce noise if samples are not representative \\
    Class Weights & Assigns higher weights to the minority class during training & No need to modify dataset, efficient & May not fully address severe imbalance \\
    \bottomrule
  \end{tabularx}
  \label{tab:class_imbalance_strategies}
\end{table}

\section{Results}
The performance of our models is evaluated using the test set, with a focus on metrics such as ROC-AUC, precision, recall, and F1-score. The 1D CNN with multi-head attention achieves an ROC-AUC score of 0.9644, demonstrating its ability to distinguish between true and false alarms with high accuracy. The precision and recall for false alarms are 0.83 and 0.85 , respectively, while the precision and recall for true alarms are 0.94 and 0.93.

These results indicate that the model performs well on both classes, with a slight bias toward detecting true alarms. The FCNN achieves even higher ROC-AUC scores across different SMOTE ratios. For example, with a SMOTE ratio of 0.5 , the model achieves an ROC-AUC score of 0.9681 . Increasing the SMOTE ratio to 0.75 improves the ROCAUC score to 0.9734 , while a SMOTE ratio of 1.0 results in a score of 0.9608 . These results suggest that balancing the dataset using SMOTE can improve model performance, particularly for the minority class.

ADASYN, another oversampling technique, achieves an ROC-AUC score of 0.9542 . While this score is slightly lower than the best-performing SMOTE configuration, it still demonstrates the effectiveness of oversampling in handling class imbalance. Finally, using class weights without data augmentation achieves an ROC-AUC score of 0.9644, highlighting the importance of addressing class imbalance during training.

The decision support system generates alerts based on a confidence threshold of 0.5. For example, if the model predicts a true alarm with high confidence, it generates an alert.

Conversely, if the confidence is low, no alert is generated. This approach ensures that healthcare providers are only notified of alarms that are likely to be true, reducing the cognitive burden of alarm fatigue.

\section{Conclusion}
This paper presents a machine learning approach to reduce false VT alarms in ICU settings using the VTaC dataset. Our results demonstrate that deep learning models, combined with effective feature extraction and class imbalance handling, can achieve high performance in distinguishing true and false alarms. The ROC-AUC scores exceeding 0.96 highlight the potential of this approach to improve the accuracy of VT alarm detection and reduce alarm fatigue in clinical settings.

By addressing the challenge of false VT alarms, this work contributes to improving patient safety and reducing the cognitive burden on healthcare providers in ICUs. The combination of advanced machine learning techniques and high quality annotated data represents a significant step forward in the fight against alarm fatigue.

\bibliographystyle{unsrt}  
\bibliography{templateArxiv}

\end{document}